\documentclass[10pt,twocolumn,letterpaper]{article}

\usepackage{wacv}
\usepackage{times}
\usepackage{microtype}
\usepackage{float}
\usepackage{epsfig}
\usepackage{graphicx}
\usepackage{amsmath}
\usepackage{amssymb}
\usepackage{multirow}
\usepackage{booktabs}
\usepackage[bookmarks=false]{hyperref}
\usepackage{enumitem}
\usepackage{tabularx}

\newenvironment{tight_itemize}{
\begin{itemize}[leftmargin=20pt]
  \setlength{\topsep}{0pt}
  \setlength{\itemsep}{2pt}
  \setlength{\parskip}{0pt}
  \setlength{\parsep}{0pt}
}{\end{itemize}}

\newcommand{\MC}[1]{\textcolor{black}{{#1}}}

\wacvfinalcopy 

\def\wacvPaperID{650} 

\ifwacvfinal  \fi

\begin{document}

\def\autonue{IDD}
\title{IDD: A Dataset for Exploring Problems of \\Autonomous Navigation in Unconstrained Environments}

\author{Girish Varma$^1$ \qquad Anbumani Subramanian$^2$ \qquad Anoop Namboodiri$^1$ \\ 
Manmohan Chandraker$^3$ \qquad C V Jawahar$^1$\\ 
$^1$IIIT Hyderabad \qquad $^2$Intel Bangalore \qquad $^3$University of California, San Diego\\ 
\url{http://idd.insaan.iiit.ac.in/}
}



\twocolumn[{%
\renewcommand\twocolumn[1][]{#1}%

\maketitle

\ifwacvfinal\thispagestyle{empty}\fi

}]

\begin{abstract}
While several datasets for autonomous navigation have become available in recent years, they tend to focus on structured driving environments. This usually corresponds to well-delineated infrastructure such as lanes, a small number of well-defined categories for traffic participants, low variation in object or background appearance and strict adherence to traffic rules. We propose \autonue, a novel dataset for road scene understanding in unstructured environments where the above assumptions are largely not satisfied. It consists of 10,004 images, finely annotated with 34 classes collected from 182 drive sequences on Indian roads. The label set is expanded in comparison to popular benchmarks such as Cityscapes, to account for new classes. 
It also reflects label distributions of road scenes significantly different from existing datasets, with most classes displaying greater within-class diversity.
Consistent with real driving behaviors, it also identifies new classes such as drivable areas besides the road. We propose a new four-level label hierarchy, which allows varying degrees of complexity and opens up possibilities for new training methods. Our empirical study provides an in-depth analysis of the label characteristics. State-of-the-art methods for semantic segmentation achieve much lower accuracies on our dataset, demonstrating its distinction compared to Cityscapes. Finally, we propose that our dataset is an ideal opportunity for new problems such as domain adaptation, few-shot learning and behavior prediction in road scenes.



\end{abstract}


\begin{figure}[!!t]
\centering
\includegraphics[width=0.48\linewidth]{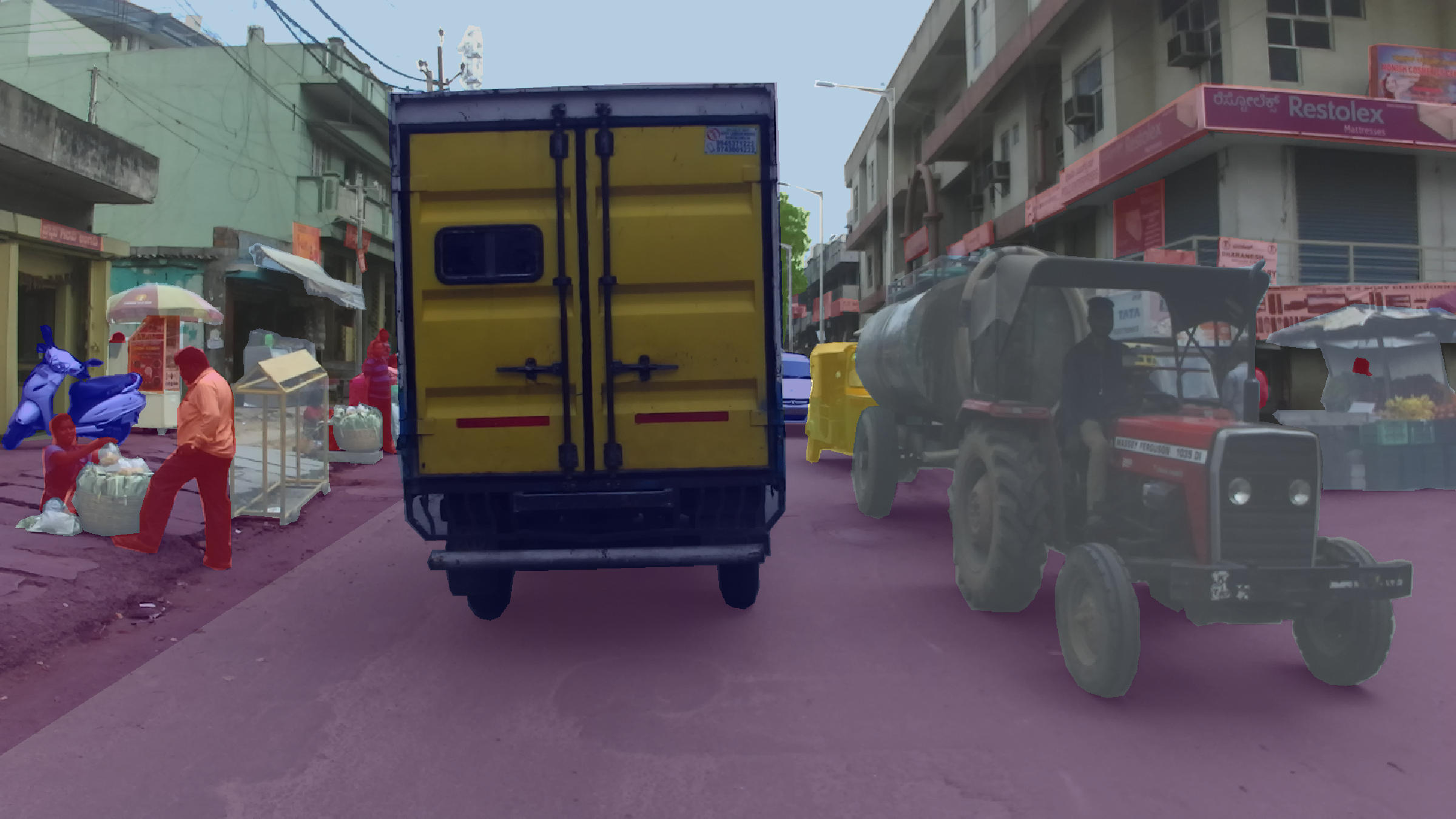} \includegraphics[width=0.48\linewidth]{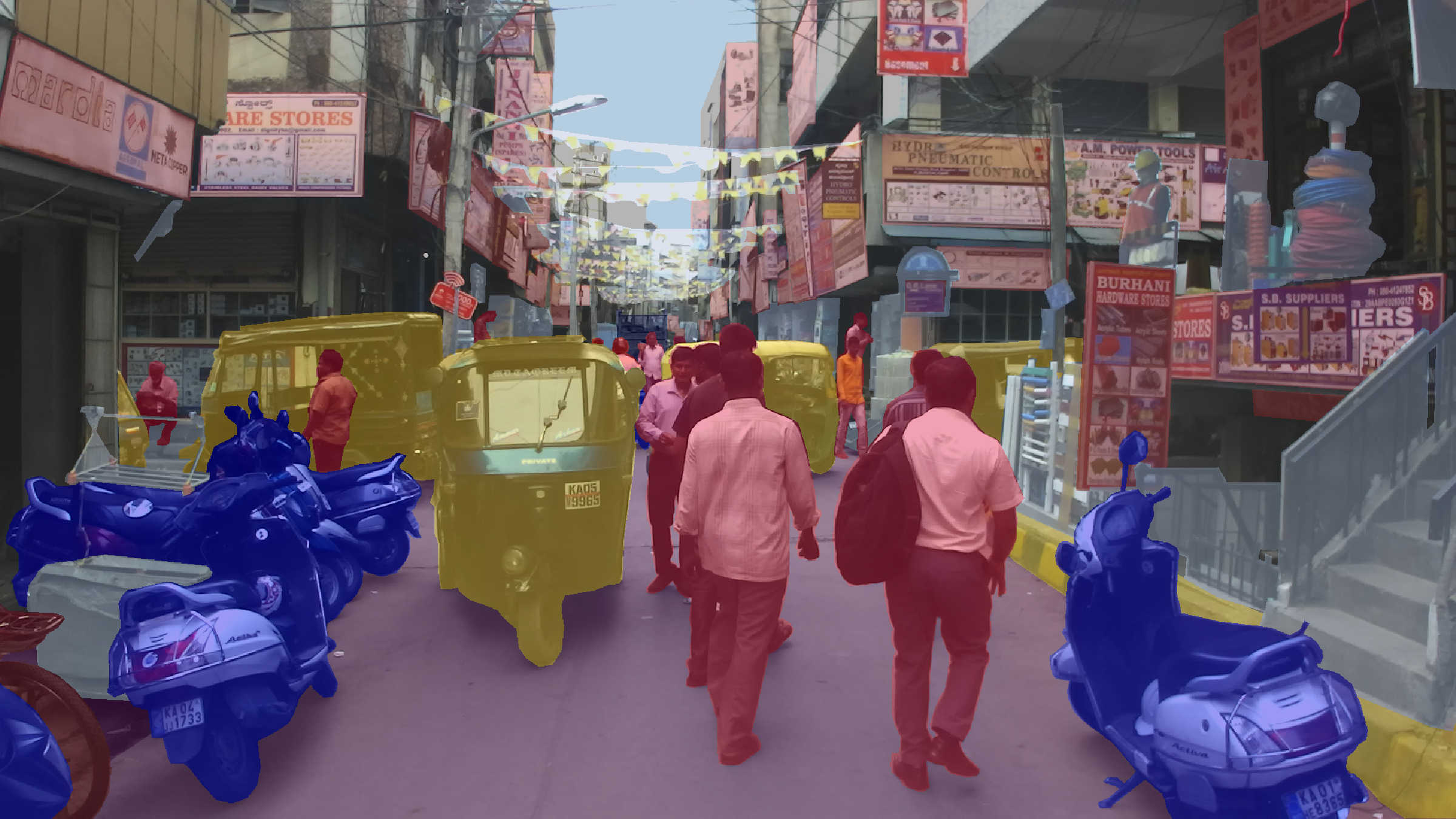}\\ \vspace{0.1em}
\includegraphics[width=0.48\linewidth]{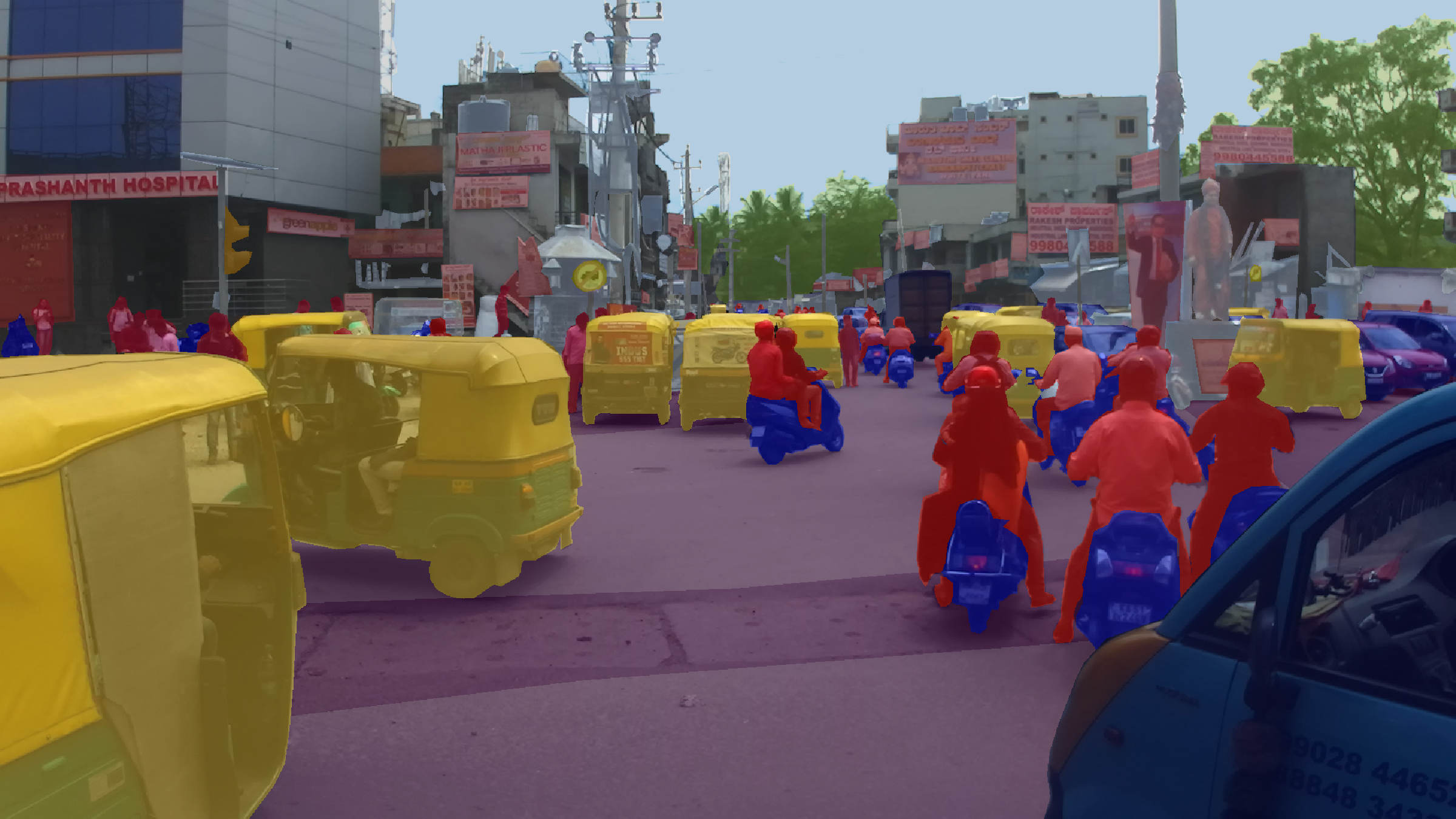} \includegraphics[width=0.48\linewidth]{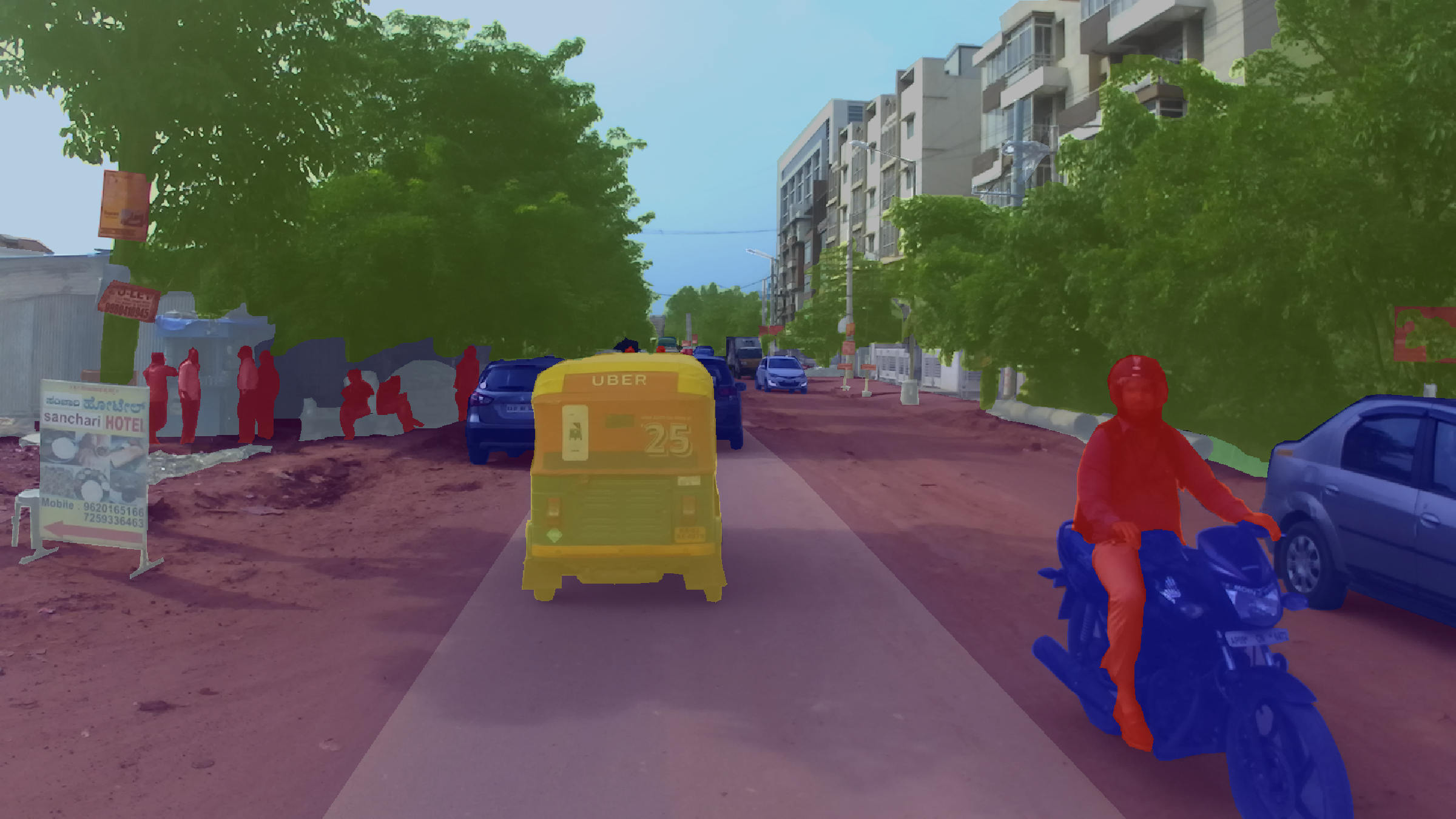}
\caption{\label{fig:examples} Some examples of the diverse and unstructured conditions that is covered by the dataset.}
\vspace{-1.5em}
\end{figure}

\section{Introduction}
\label{sec:intro}

Autonomous navigation is rapidly maturing towards becoming a mainstream technology, with even consumer deployment by major automobile manufacturers. A significant contributor to this progress has been the availability of large-scale datasets for sensing and scene understanding. Yet, several challenges remain in enabling self-driving across diverse geographies. A key challenge is to achieve data scale and diversity large enough to ensure safety and reliability in extreme corner cases. Even more importantly, algorithms are largely untested in their ability to generalize to road conditions that are significantly more diverse and unstructured.

In this paper, we propose IDD, a dataset that takes the first steps towards addressing the above concerns. Our dataset shares several traits such as scale, annotation and tasks with similar ones in structured environments, namely KITTI \cite{KITTI} or Cityscapes \cite{CS}. But it also intends to significantly expand the scope of the autonomous navigation problem, along each of those dimensions.
Similar to Cityscapes \cite{CS}, we provide large-scale raw data with multiple cameras and sensors across cities and lighting conditions. But the scale of our data is larger, consisting of $10,004$ labeled images with fine instance-level boundaries. Next, while the annotation type is similar for our dataset and Cityscapes \cite{CS}, the number of object classes and within-class diversity of appearance are higher for IDD. Finally, while we also initially propose instance segmentation as the task of interest, our label diversity and novel hierarchy might allow novel machine learning techniques and computer vision algorithms.

\begin{figure*}[!th]
    \vspace{-4em}
    \centering
    \includegraphics[width=1.\textwidth]{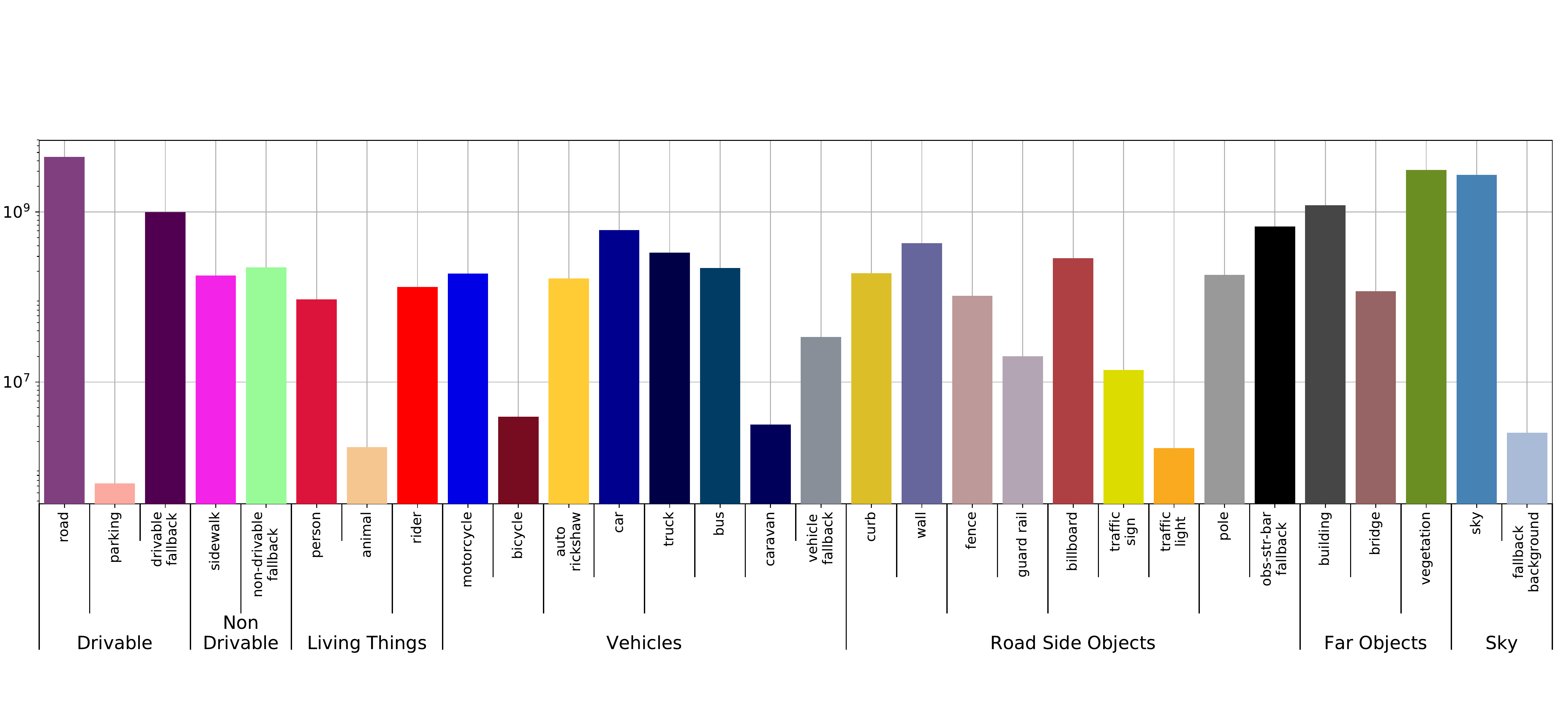}
    \vspace{-3em}
    \caption{\label{fig:label-heirarchy} Label distribution in our dataset. The following information is shown here: (i) pixel counts of individual labels on the y-axis (ii) four-level label hierarchy used by the dataset at the bottom, (iv) the color legend for the predicted and ground truth masks shown in the paper is used for the corresponding bars.  We define metrics at 4 levels of the hierarchy with 30 (level 4), 26 (level 3), 16 (level 2) and 7 (level 1) labels, respectively, giving different complexity levels for training models. }
    \vspace{-1em}
\end{figure*}

The singular defining aspect of our dataset is that it corresponds to driving in less structured environments. We argue that this is a better reflection of the needs for autonomous navigation in large portions of the world, including Asia, South America and Africa. Accordingly, we collect our data in India where road scenes differ markedly from those in Europe or North America. The variety of traffic participants in Indian roads is larger, including novel classes such as autorickshaws or animals. The within-class diversity is also higher, for example, since vehicles span a larger range of manufacturing years and ply with larger variation in wear. Even the distribution of classes that overlap with Cityscapes is significantly different, for instance, the proportion of motorcycles is far higher, as is that of multiple riders on two-wheeled vehicles. Background classes also display greater diversity, such as city scenes rich in novel classes such as billboards. Besides variations in weather and lighting, other ambient factors such as air quality and dust also span greater ranges in our dataset. Such greater complexity of road scenes necessitates a larger scale of data. Thus, we provide high-quality annotation at a scale significantly larger than available for other contemporary datasets such as KITTI or Cityscapes.
\footnote{Datasets such as Berkeley Deep Driving \cite{BDD100K} and Apolloscape \cite{apolloscape}  have recently been released with labels at a similar scale. However, they are contemporaneous with our work, which precludes a detailed comparison. In any case, we note that they are in structured environments, which makes our dataset clearly different.}

We provide a detailed analysis of the label distributions in the IDD dataset, while highlighting some of the above differences. We also showcase those differences through quantitative evaluation of state-of-the-art algorithms on our dataset. We consistently observe that semantic segmentation performances are far lower on IDD as compared to Cityscapes, using identical models and with larger-scale training data for IDD. Firstly, this highlights that conventional semantic segmentation datasets such as Cityscapes are getting saturated and the next set of challenges lie in more complex datasets like IDD. Secondly, this highlights the need for ever-larger training data as we expand the scope of the autonomous navigation problem to newer geographies.

Besides segmentation, the nature of our dataset also enables novel problems for vision and learning. This is already reflected in some of our annotation choices. For instance, while the notion of a drivable area in Europe is largely defined by classes such as roads or lanes that have distinct appearances, it is more ambiguous in our dataset and likely also informed by semantic cues such as presence of dynamic traffic participants. Thus, we include labels for safely drivable and non-drivable areas. Our label hierarchy is attuned to the autonomous navigation problem and we postulate that exploiting it might lead to semantic segmentation more suited to subsequent applications such as collision avoidance or path planning. We label classes that are rare but important for navigation (such as animals), or classes that exhibit large within-class variance (such as autorickshaw), which motivates problems such as few-shot learning.

The contrast of our dataset with structured ones also suggests interesting directions of future research. For instance, domain adaptation between Cityscapes and IDD is clearly a need given the large performance drops encountered in cross-dataset settings. Classes that are unique to our dataset also encourage consideration of domain adaptation with non-overlapping label spaces. Even higher-level reasoning problems such as behavior prediction pose new challenges in IDD, since traffic participants have lower adherence to traffic rules, motions can be sudden, complex obstructions might be present, drivable areas can be ambiguous and traffic lanes need not correspond to lane markings on the road. While not considered in this paper, we highlight that these novel problems do arise in unstructured environments such as ours.

\section{Challenges in Unstructured Environments}
\label{sec:challenges-india}
We collect data from Indian roads and analyze the shortcomings of models trained on existing datasets. As illustration, we describe some of the qualitative issues observed when using predicted outputs of a model that obtains 70\% mean IoU on the Cityscapes validation set. 

\begin{figure}[!!t]
\centering
\includegraphics[width=0.49\textwidth]{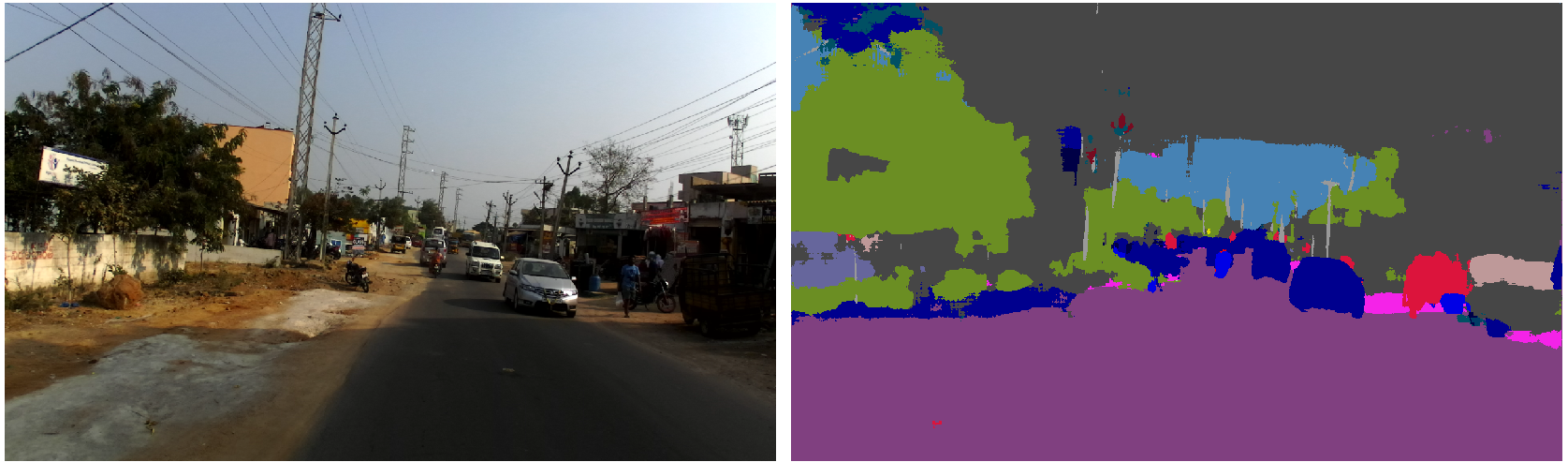}
\includegraphics[width=0.49\textwidth]{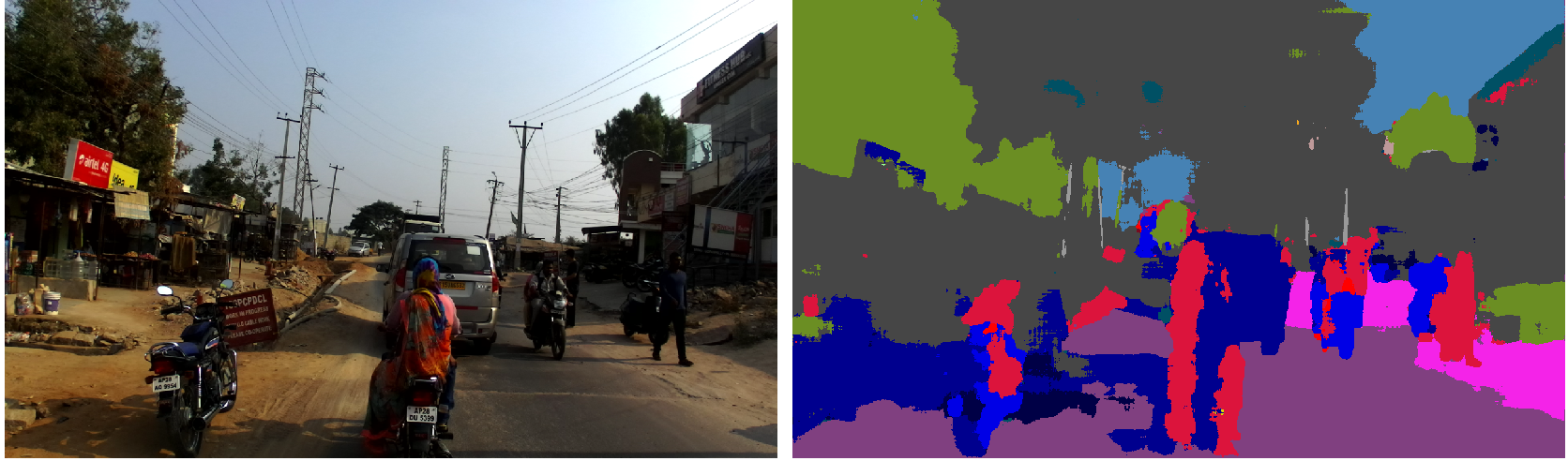}
\vspace{-1em}
\caption{\label{fig:roadside} Cityscape models do not distinguish between the road and possible unsafe drivable area on both sides of the road.}
\vspace{-1em}
\end{figure}

\vspace{-0.3cm}
\paragraph{Ambiguous Road Boundaries.}
Road boundaries in Cityscapes are very well defined and usually flanked on both sides by barriers or sidewalks. However, this is not the case in our setting. Road sides can have muddy terrain, while also being drivable to some extent. Roads themselves can be covered by dirt or mud, making the boundaries very ambiguous. On the other hand, Cityscapes models often recognize flat areas beside the road which need not be safe for driving as road, as  seen in Figure \ref{fig:roadside}.

\vspace{-0.3cm}
\paragraph{Diversity of Vehicles and Pedestrians.}
Indian roads have a variety of unique vehicles like auto-rickshaws, which behave very differently than other vehicles like cars. \MC{Even for standard categories like cars, the appearance variations are higher due to greater wear and tear.} Further, the frequency and variety of trucks and buses are also high. Another distinction is the large number of motorbikes with multiple persons riding it. Pedestrians often cross the road at arbitrary locations, rather than crosswalks. Bikes and autorickshaws \MC{are also less likely to follow traffic discipline, thus, there are fewer correlation between traffic participants and road signage such as lanes or traffic lights.}

\begin{figure}
\centering
\includegraphics[width=0.49\textwidth]{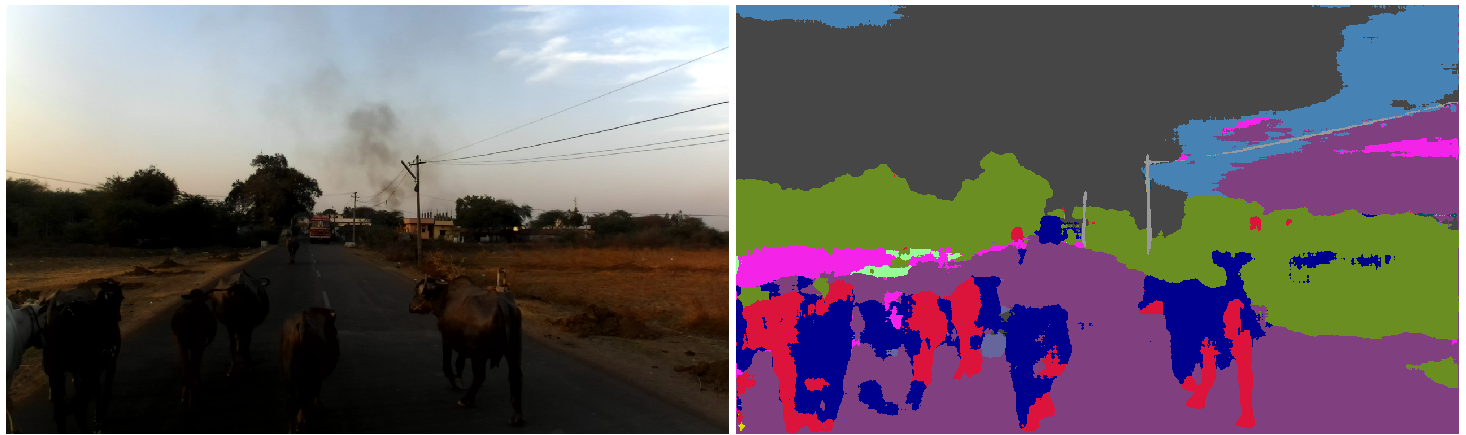}
\includegraphics[width=0.49\textwidth]{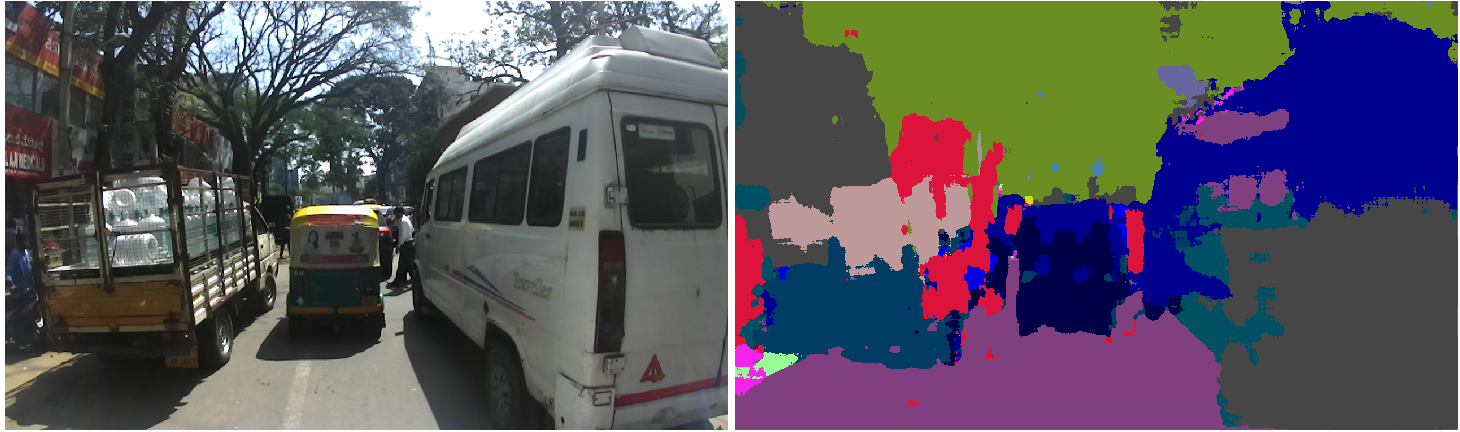}
\includegraphics[width=0.49\textwidth]{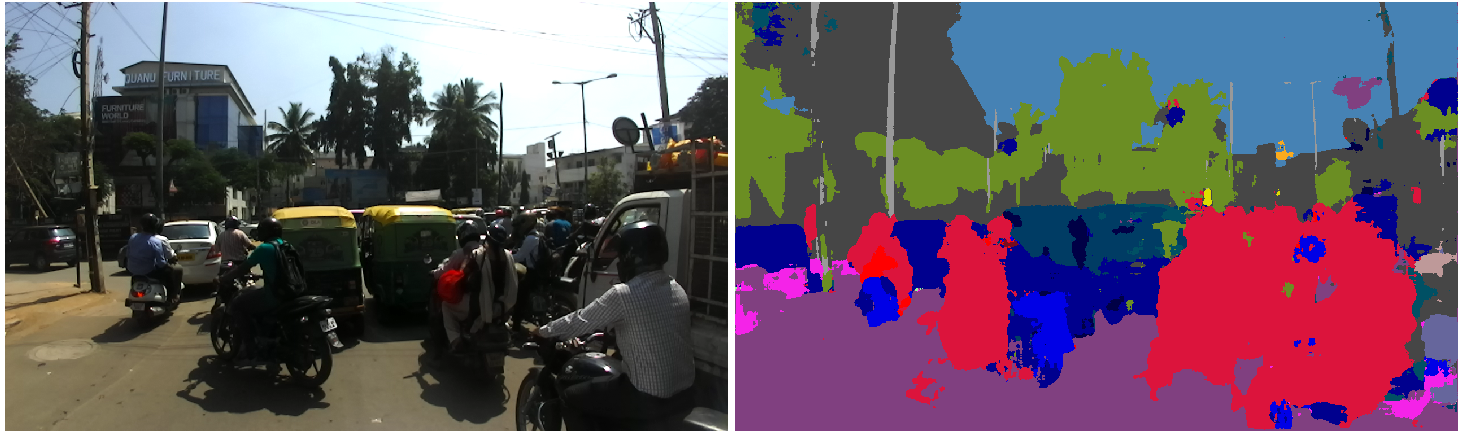}
\includegraphics[width=0.49\textwidth]{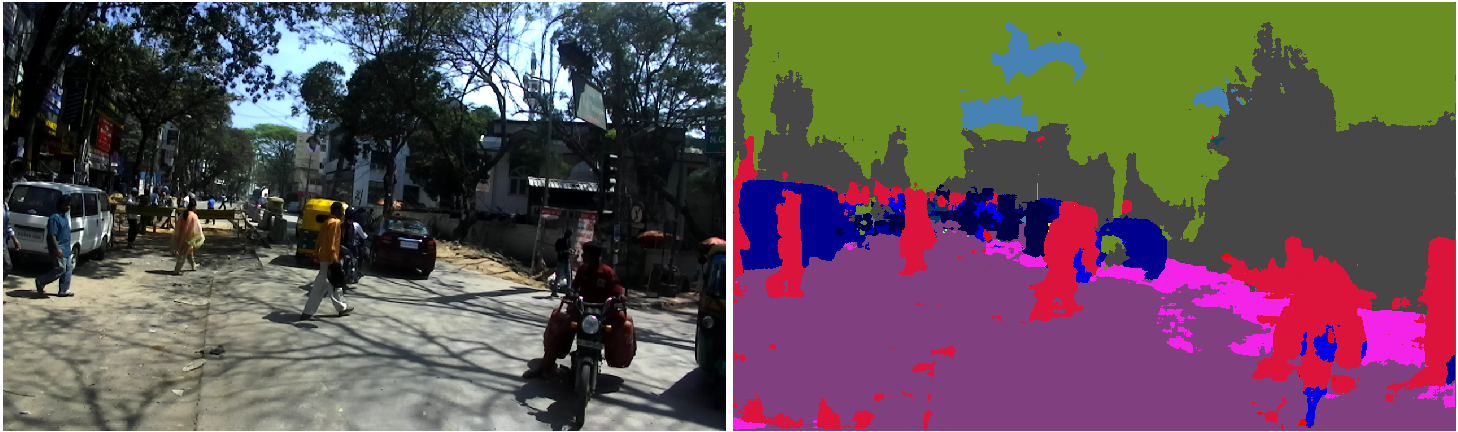}
\vspace{-1em}
\caption{\label{fig:div-vehicles} (Top) A herd of buffaloes on the road at dusk. (Bottom) Many motorbikes with multiple riders, not necessarily following traffic rules such as road lanes.}
\vspace{-2em}
\end{figure}

\vspace{-0.3cm}
\paragraph{Extensive Use of Information Boards.}
Information displays such as billboards appear extensively in our dataset. They can be significant for localization and mapping problems by indicating buildings or landmarks. Sometimes they also indicate special vehicles, such as advertisements attached to a driving school car, or a delivery vehicle.

\begin{figure}
\centering
\includegraphics[width=0.49\textwidth]{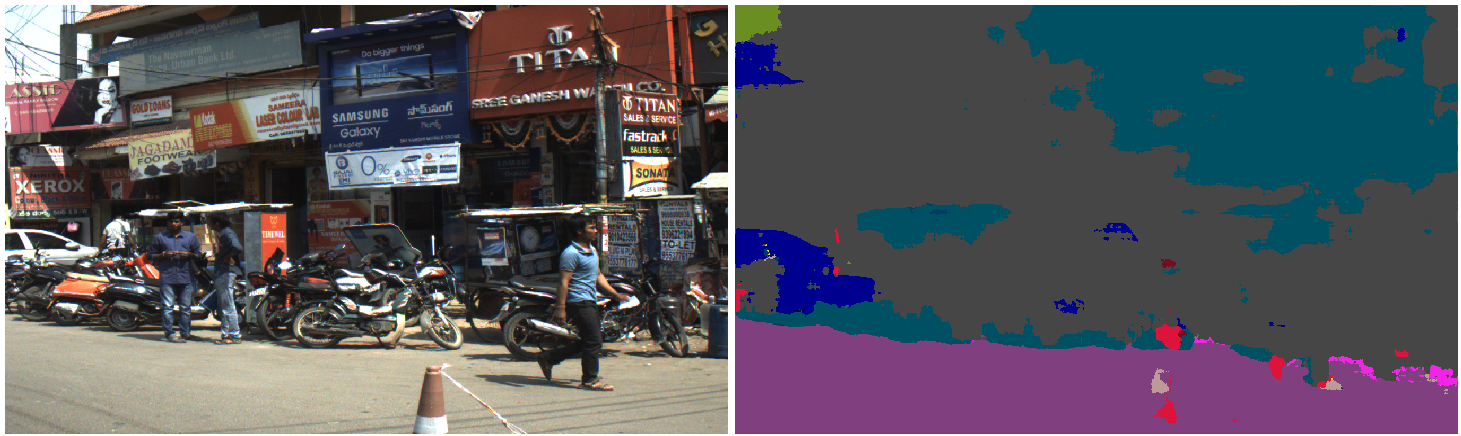}
\includegraphics[width=0.49\textwidth]{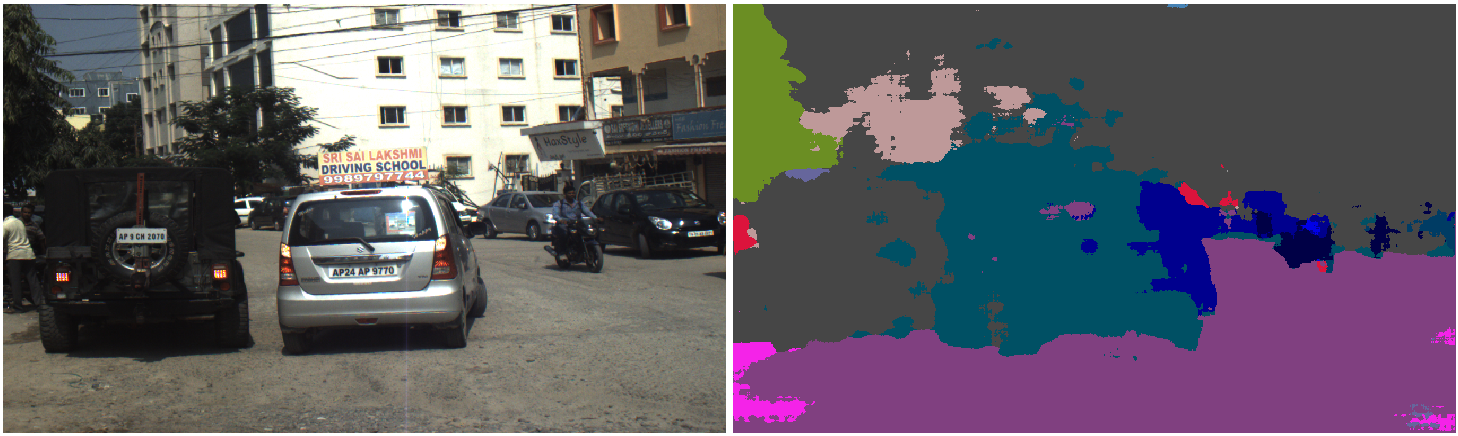}
\vspace{-1em}
\caption{\label{fig:billboards} (Left) An array of billboards indicating the shops. (Right) A vehicle with a billboard of a driving school.}
\vspace{-2em}
\end{figure}

\begin{figure}
\centering
\includegraphics[width=0.49\textwidth]{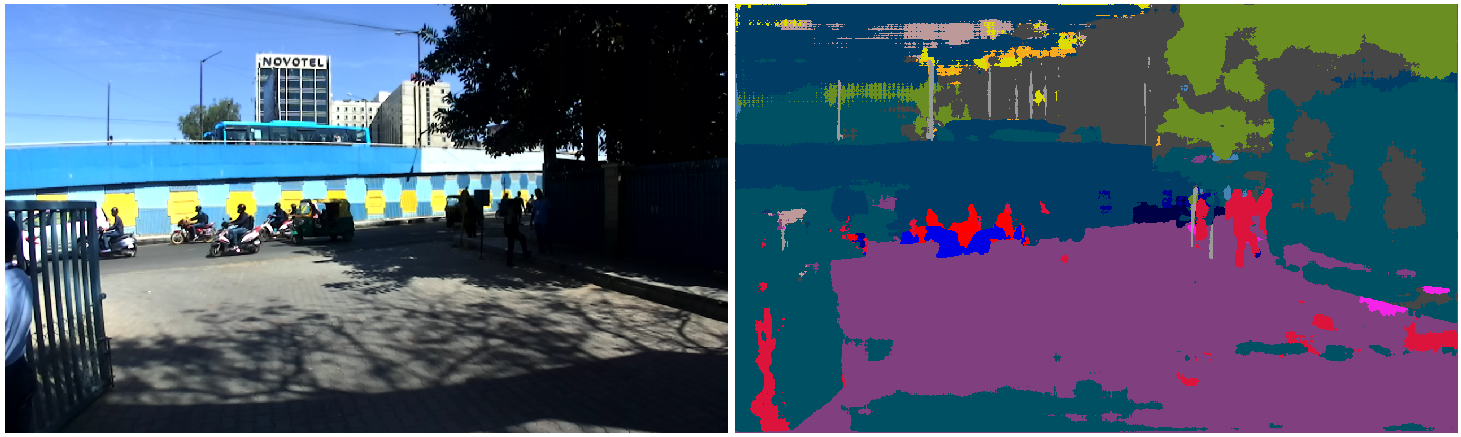}
\includegraphics[width=0.49\textwidth]{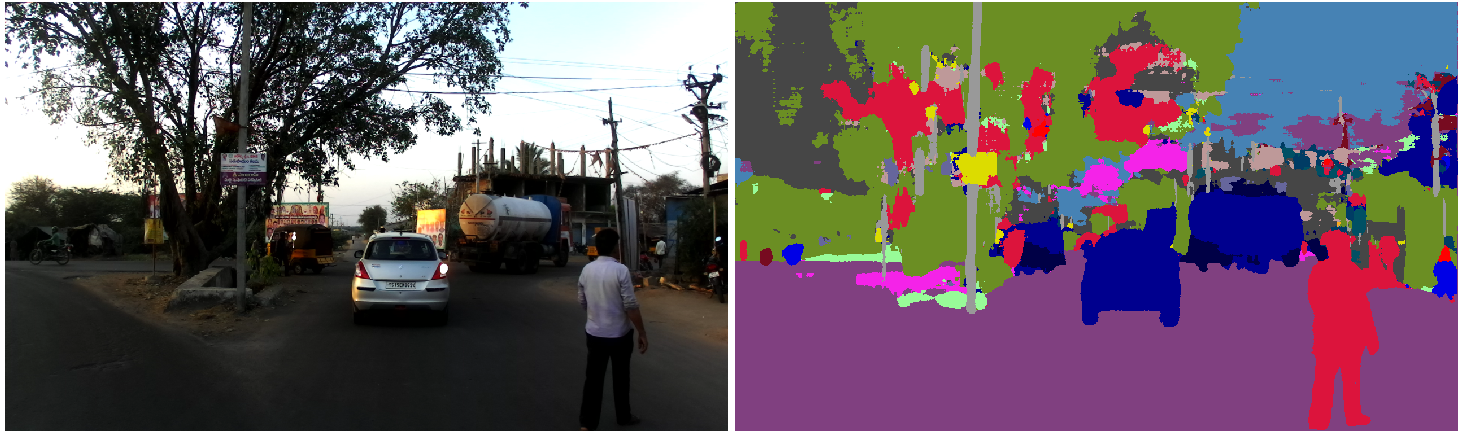}
\vspace{-1em}
\caption{\label{fig:shade1} Heavy shadows in the image (top) or low light conditions (bottom) can greatly degrade the quality of predictions using models trained on Cityscapes. }
\vspace{-1em}
\end{figure}

\vspace{-0.3cm}
\paragraph{\MC{Diversity of Ambient Conditions.}}
\MC{Lighting variation in our dataset is high since we acquire images at various times of the day, including mid-day, dawn and dusk.}
Also, some of the images have heavy shadows, which are common during a long summer season. Our dataset also contains scenes with heavily clouded skies. \MC{The greater variation in particulate matter due to fog, dust or smog also leads to significant appearance variations.} Cityscapes pretrained models yield lower accuracies in these settings, as seen in Figure \ref{fig:shade1}.

\section{Dataset}\label{sec:spec}

\begin{table*}
  \centering

    \begin{tabular}{l l l l l l l l l r }
    \toprule  \\[-1em]
   Dataset  & Calibration & \parbox{1cm}{Nearby\\frames\\ / Video} & \parbox{1.5cm}{Distortion\\ /Night}  &  \parbox{1.5cm}{\#Images/ \\ \#Sequences} & \parbox{1.5cm}{\#Labels\\Train/Total} & \parbox{1.5cm}{Average\\ Resolution}   \\[1.2em]
      \midrule \\[-2ex]
   Cityscapes \cite{CS}  & \checkmark &\checkmark & & 5K / ~50 & 19/34 & 2048x1024    \\
   \autonue    & \checkmark & \checkmark & &  10K / 180 &  30/34  & 1678x968  \\
   BDD100K \cite{BDD100K}   &  & \checkmark & \checkmark  & 10K / 10K & 19/30 & 1280x720   \\
   MVD \cite{Mappilary}  & &  &  & 25K / ~-~~ & 65/66 & $>$1920x1080   \\
      \bottomrule \\[-1.5ex]
    \end{tabular}
    \caption{\label{tab:ds-comp} Comparison of semantic segmentation datasets for autonomous navigation.}
    
    \vspace{-1.2em}
  \end{table*}

\subsection{Acquisition}

The data was collected from Bangalore and Hyderabad cities \MC{in India} and their outskirts. The locations have a mix of urban and rural areas, highway, single lane and double lane roads with a variety of traffic. The driving conditions in these localities are highly unstructured due to multiple reasons: (i) these cities are rapidly growing and have a lot of construction near the roads, (ii) road boundaries are not well defined, (iii) pedestrians and jaywalkers are aplenty in these road images, and (iv) high density of  motorbikes and trucks on the road. The variety of vehicle models are also very large. A total of 182 drive sequences were used for the preparation of the dataset.


\subsection{Frame Selection}
 We chose images from one of the forward facing cameras of a stereo pair, for  fine annotation. Images were sampled at varying rates from the video sequence, with denser sampling around crowded and special interest places like traffic junctions. These images were annotated very finely, by layered polygon masks similar to Cityscapes. Since the road conditions are highly unstructured, we need a wider variety of labels. We annotated a total of 10,004 frames.

\begin{figure*}
  \centering
  \includegraphics[width=\textwidth]{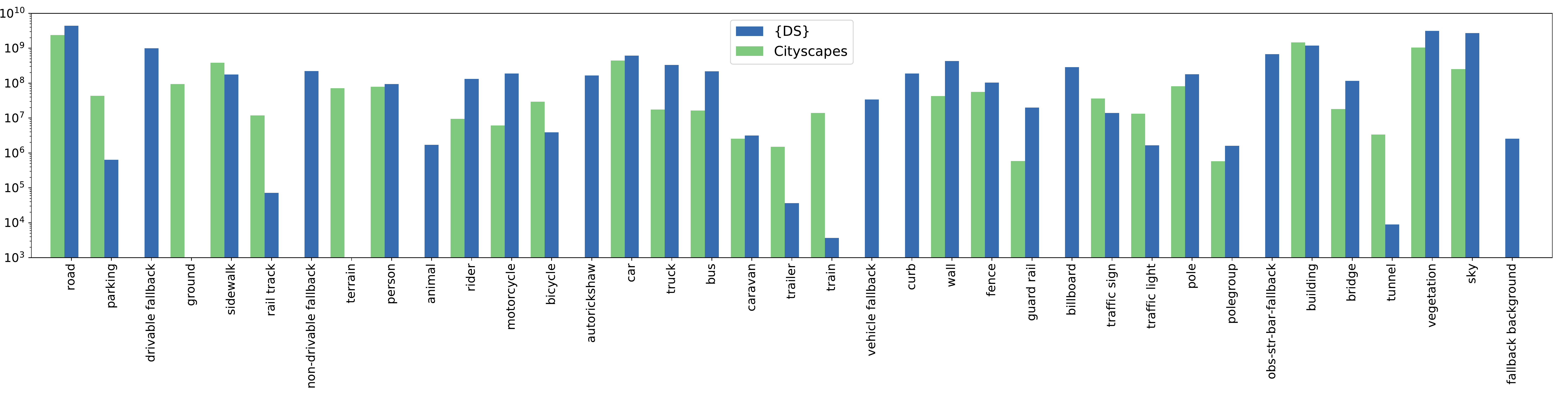}
  \vspace{-1em}
  \caption{\label{fig:pix_comp_cs} Comparison of the pixel count in our dataset with Cityscapes. The y axis is plotted in log-scale. Note that for most classes of vehicles, the number of pixels are 5-10 times more than Cityscapes. Moreover our dataset has newer labels like autorickshaw, billboard, drivable/nondrivable fallback which also have significant number of labeled pixels.}
  \vspace{-1.2em}
  \end{figure*}

\subsection{Label Hierarchy \& Annotation}
We used a total of 34 labels in the fine annotations. The labels were given definition by means of a textual description as well as example images. However, we found that it is difficult to completely avoid ambiguity between some labels. For example labels like parking, caravan or trailer cannot be precisely defined due to the diversity of the scenes and vehicles in the data collected. For resolving this issue, we designed a 4 level label hierarchy having 7 (level 1), 16 (level 2), 26 (level 3) and 30 (level 4) labels (see Figure \ref{fig:label-heirarchy}). Each level defines a category as the union of labels in the succeeding level, which are chosen such that they are ambiguous. Since we take unions of the most ambiguous labels while designing the hierarchy, the lower levels have lesser ambiguity. We have a set of new labels not available in Cityscapes \cite{CS} like auto rickshaw, billboards, animal, curb. We also have separate labels for road, drivable fall-back and non-drivable fall-back indicating safe, unsafe and non-drivable flat surfaces. We have added fall-back labels whenever appropriate so that highly ambiguous objects can be given labels.

For labeling the dataset, the annotation team was first asked to re-annotate images from the Cityscapes \cite{CS} dataset. The difference between the annotations were subsequently shown to the annotators. This process was done until the annotators were achieving greater than $95$\% accuracy with respect to the Cityscapes ground truth labels.

\subsection{Statistical Analysis and Dataset Splits}
The pixel statistics among the labels can be seen in Figure \ref{fig:label-heirarchy}. The labels in level 4 have high class imbalance. Labels like parking, animal, caravan or traffic light have much fewer pixels. The annotated dataset also has labels for trailer and rail track, which were combined with vehicle fallback and nondrivable fallback in level 4, since they have very few pixels that mostly fell within a few drive sequences. The lower levels are designed such that the imbalance is lesser.

Class imbalance at level 4, creates a problem while splitting the dataset in train, test and validation sets. Since the splitting is done at the level of drive sequences (that is, all images within a drive sequence are moved to the same split), we need to be careful that the few drive sequences that contains a label are rightly split. We roughly divide the dataset in to 70\% train, 10\% validation and 20\% test splits. The split was done by randomly assigning the drive sequences with the said distribution. We did the splitting multiple times to come up with a split where all the 30 labels in level 4 have approximately 70, 10 and 20 percentage of pixels in  the train, validation and test sets, respectively.

\subsection{Comparison with Other Datasets}
\begin{figure}
  \centering
  \includegraphics[width=0.45\textwidth]{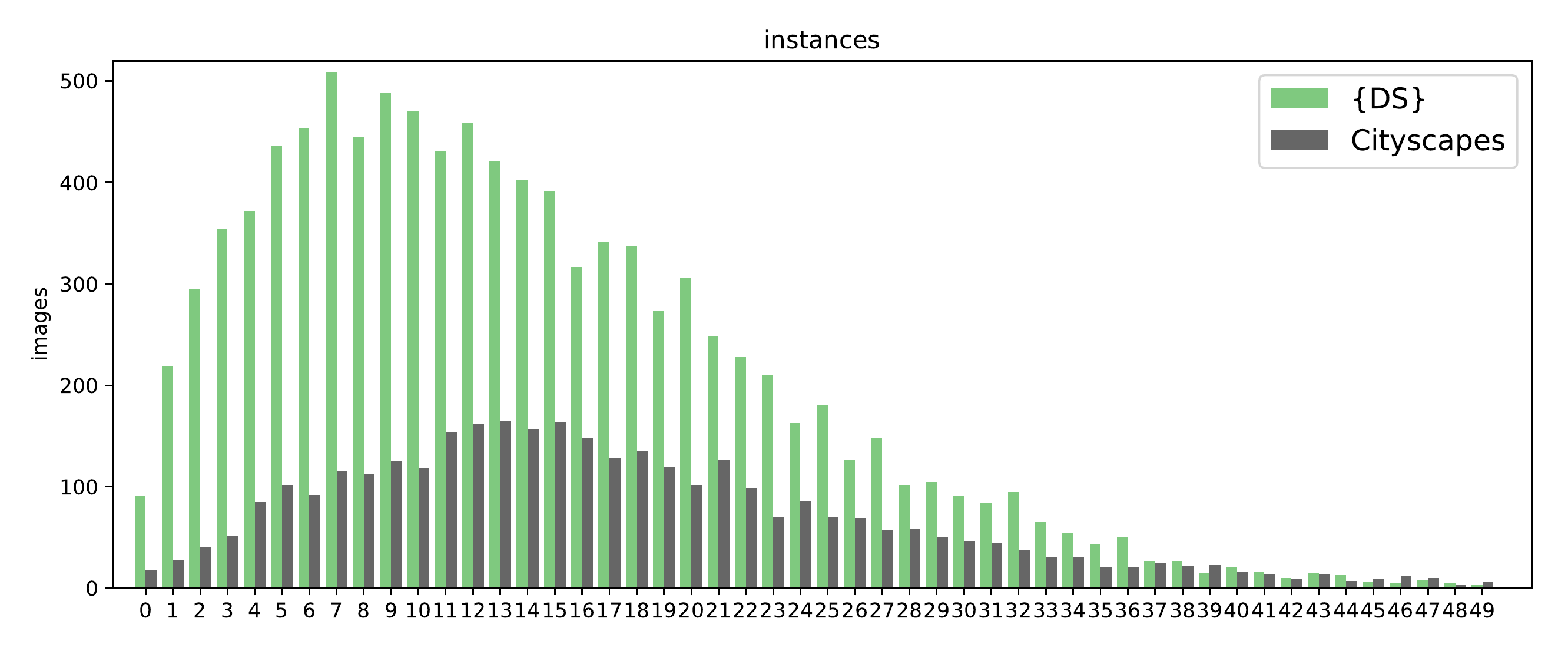}  
  \caption{\label{fig:inst_tp} Comparison of traffic participants in our dataset with Cityscapes.}
  \vspace{-1.5em}
\end{figure}
Various datasets have been proposed for studying the semantic and instance segmentation problems like Pascal VOC \cite{pascal}, MS COCO \cite{coco}, SUN \cite{sun,sun2}.
The datasets proposed for semantic segmentation that focus on autonomous navigation are Cityscapes \cite{CS}, KITTI \cite{KITTI}, Camvid \cite{camvid,camvid2}, Leuven \cite{leuven} and the Daimler Dataset \cite{dailmerdataset}. ADE20K \cite{ADE20K} is a recent dataset which focuses on the general scene parsing problem. More recently Mapillary Vistas dataset \cite{Mappilary} (which focuses on street view imagery) and the Berkeley Deep Drive Dataset \cite{BDD100K} (for autonomous navigation) was released. A comparison of the metadata available in these datasets can be seen in Table \ref{tab:ds-comp}. As can be seen, our dataset is more similar to Cityscape in the sense that we collect data from calibrated cameras without any distortions. BDD100K uses dashboard cameras kept inside the car and hence often has internal reflections from the glass as well as rain distortions. Moreover a good fraction of the dataset consists of night images. Mapillary Vistas dataset consists of images taken using a variety of cameras (including smart phones) having varying perspectives of the road and the road side. They do not have video data or images of near by frames.

\begin{table}
  \centering
  \begin{tabular}{l c c c}
      \toprule   \\[-1em]
      \multirow{2}{*}{\parbox{2.5cm}{Method}} &
        \multicolumn{3}{c}{\parbox{4cm}{\centering \% mIoU at {Hierarchy Levels}}}  \\
        \cmidrule{2-4} 
      & \parbox{1.2cm}{\centering 1} &  \parbox{1.2cm}{\centering 2} &  \parbox{1.2cm}{\centering 3} \\
      \midrule \\[-2ex]
      
      GT subsampled by 2 & 99 & 97 & 97\\
     
      GT subsampled by 4 & 98 & 96  & 95\\
      
      GT subsampled by 8 & 96 & 92  & 90\\
      
      GT subsampled by 16 & 92 & 87  & 84\\
      
      GT subsampled by 32 & 86 & 78  & 74\\
      
      GT subsampled by 64 & 77 & 66  & 61\\
      
      GT subsampled by 128 & 65 & 53  & 48\\
      
      \bottomrule \\[-1.5ex]
    \end{tabular}
    \caption{\label{tab:control-expts}  Control  experiments  to  estimate  upper  bounds  for  se-
    mantic segmentation results, assessed by Intersection-over-Union
    (IoU, in \%) scores for different levels of the hierarchy.}
    \vspace{-1em}
  \end{table}
  
We compare the label statistics of our dataset with Cityscapes (since it is more similar to our dataset as described above) in terms of pixel counts (Figure \ref{fig:pix_comp_cs}) and the number of instances of traffic participants (Figure \ref{fig:inst_tp}). We have more pixels of truck, bus, motorcycle, guard rail, bridge and rider (see Figure \ref{fig:pix_comp_cs}). The pixel counts for new labels (auto rickshaw, billboard, curb, drivable-fallback, nondrivable-fallback) are also high. In terms of instances of traffic participants, we have almost double the counts, with a distribution similar to Cityscapes (see Figure \ref{fig:inst_tp}).

\section{Benchmarks}\label{sec:benchmarking}

\begin{table*}[!tbh]
  \centering
  
  \begin{tabular}{ l  l l  l*{16}{c}  l } 
  \toprule   \\[-1em] 
  \rotatebox{90}{\scriptsize Train} & \rotatebox{90}{\scriptsize Test} & \rotatebox{90}{\scriptsize road} & \rotatebox{90}{\scriptsize sidewalk} & \rotatebox{90}{\scriptsize person}  & \rotatebox{90}{\scriptsize motorcycle} & \rotatebox{90}{\scriptsize bicycle} & \rotatebox{90}{\scriptsize car} & \rotatebox{90}{\scriptsize truck} & \rotatebox{90}{\scriptsize bus} & \rotatebox{90}{\scriptsize wall} & \rotatebox{90}{\scriptsize fence} & \rotatebox{90}{\scriptsize traffic sign} & \rotatebox{90}{\scriptsize traffic light} & \rotatebox{90}{\scriptsize pole} & \rotatebox{90}{\scriptsize building} & \rotatebox{90}{\scriptsize vegetation} &  \rotatebox{90}{\scriptsize sky} & \rotatebox{90}{\parbox{1cm}{\scriptsize mIoU of\\[-0.2em] common\\[-0.2em] labels}} \\
  \midrule \\[-2ex] 
  CS & DS & 72 & 22 & 30  & 47& 10&  58 & 30 & 19 & 17 & 13 &  19 & 8 & 23& 32& 76& 68& 34   \\
  DS & CS & 81 & 26 & 74  &  34 & 55 & 85 & 16 & 17 & 21 & 24 & 25 & 21 & 47& 77& 90& 88 & 49   \\
  \midrule \\[-2ex]
  BD & ID & 83 & 0 &  38 & 44 & 2 & 52 & 21 & 13 & 0 & 0 & 0 & 0 & 36 & 42 & 83 & 94 & 32   \\ 
  ID & BD & 84 & 16 &  57 & 34 & 44 & 77 & 14 & 24 & 10 & 33 & 18 & 13 & 41 & 68 & 82 & 87 & 44  \\ 
  \midrule \\[-2ex]
  CS & CS & 98 & 84 & 81 & 60 & 76 & 94 & 56 & 78 & 49 & 58 & 77 & 67 & 62& 92& 92& 94& 76 \\
  MV & MV & 85 & 58 & 73 & 55 & 61 & 90 & 61 & 65 & 45 & 58 & 72 & 67 & 50 & 86 & 90 & 98 & 70 \\ 
  ID & ID & 92 & 68 & 73 & 80 & 42 & 89 & 79 & 78 & 64 & 45 & 60 & 38 & 58 & 75 & 90 & 97 & 70 \\ 
  BD & BD & 95 &  62 & 61  & 32  & 22  & 90 & 52 & 57 & 25 & 45 & 52 & 58 & 49 & 85 & 87 & 97 & 60 \\ 
  
  \bottomrule \\[-1.5ex]
  
  \end{tabular}
  \caption{\label{tab:dom-desc} The domain discrepancy between Cityscapes (CS) \cite{CS}, Mapillary Vistas (MV) \cite{Mappilary}, Berkeley Deepdrive (BD) \cite{BDD100K} Dataset and \autonue~ (ID) using the DRN-D-38 Model \cite{drn}. Performance for only the common labels between the four datasets are used. First two rows compares the accuracy of a model trained on one of  \autonue~ or Cityscapes and tested on the other dataset. As can be seen, \autonue~ trained model can predict CS and BD labels, better than predictions of trained models of the corresponding datasets on \autonue. The bottom four rows gives the performance of models in each of the datasets. \autonue~dataset is harder than CS dataset and similar in hardness to MV on these 16 labels. BD is harder because i.) it has night scenes ii.) the images are take from a dash board cam, hence has reflections from inside the car as well as distortions like rain drops on the mirror. }
  \vspace{-1.5em}
  \end{table*}

\subsection{Control Experiments}
In Table \ref{tab:control-expts}, we provide the results of some control experiments which provide upper bounds for IoU scores for models giving predictions at a given factor of the input resolution. We first  downsample  the ground truth by a given factor and then upsample it to the original image size for evaluation of average IoU at original scale. We provide the mean IoU scores of different levels of the hierarchy,  confirming  that low-resolution  processing contributes significantly to overall degradation of segmentation results.

\subsection{Domain Discrepancy}
Domain discrepancy studies the quantitative shift in data distributions between datasets. To understand it, we train a DRN-D-$38$ (Dilated Residual Networks \cite{drn}) model in Cityscapes \cite{CS}, Mapillary \cite{Mappilary}, BDD100K \cite{BDD100K} and our dataset. We compare the IoU scores in a set of 16 common labels between the four datasets in Table \ref{tab:dom-desc}. As seen from the last four rows, our dataset is harder than Cityscapes while having a similar level of hardness compared to the Mapillary Vistas Dataset. We also report IoU scores of predictions given by models trained on one dataset and tested on the other. A pretrained model trained on our dataset performs better when tested on Cityscapes and BDD100K, as compared to the converse experiment.

\subsection{Semantic Segmentation Benchmark}

\begin{figure*}
  \centering
  \includegraphics[width=\textwidth]{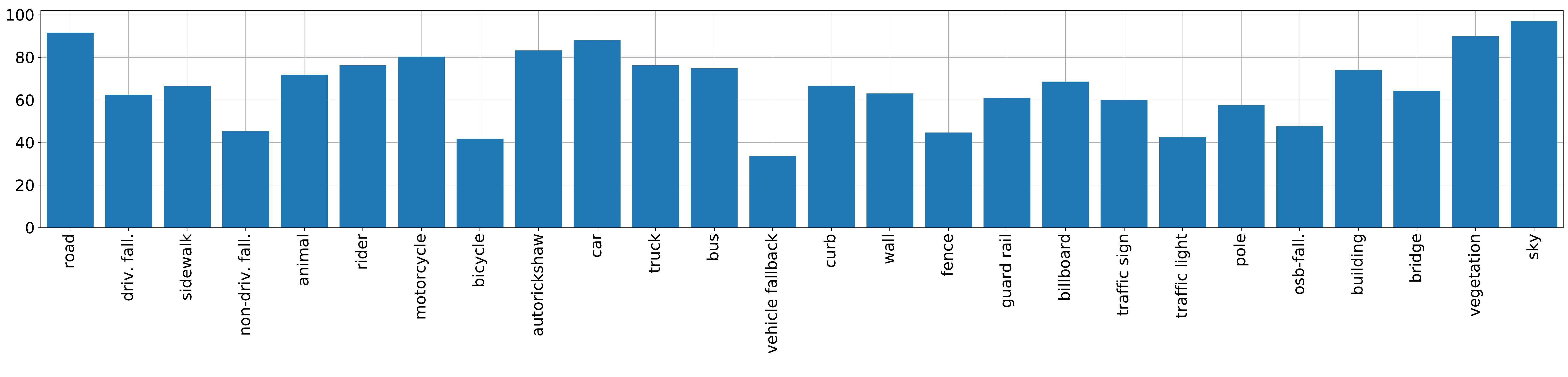} 
  \vspace{-1.8em}
  \caption{\label{fig:ious} The IoUs for every class for the DRN D 38 model trained on \autonue with mIoU of 66.5\%.}
  \vspace{-1.3em}
  \end{figure*}
The semantic segmentation benchmark on our dataset quantifies the mean Intersection over Union (mIoU) scores at the four levels of the hierarchy. There are some labels in level 4 like traffic light, parking or animal for which the number of labeled pixels are very few. Hence, this serves as an excellent benchmark for transfer learning or domain adaptation problems. We also have level 1 and level 2 mIoU scores, which are a useful benchmark for real-time models, since they might not be able give good results on the fine grained classification task at level 3 and 4. The level 1 benchmark still has classes for \MC{most of} the essential labels for autonomous navigation.

We benchmarked our dataset using the DRN-D-38 (\cite{drn}) and the ERFNet (real-time model \cite{ERFNet}) model. We also conducted a challenge and evaluated submissions which use some of the state-of-the-art models. The results are shown in Table \ref{tab:sem-bench}.

\begin{table}
  \centering
  \begin{tabular}{l c c c }
    \toprule  \\[-1em]
    \multirow{2}{*}{Method} &
      \multicolumn{3}{c}{\parbox{2.5cm}{\centering \% mIoU at Levels} }  \\
      \cmidrule{2-4} 
    & L1 &  L2 & L3  \\
    \midrule \\[-1.5ex]
    ERFNet & - & - & 55.4  \\[0.8em]
    
    DRN-D-38 & 85.9 & 72.6  & 66.6  \\[0.8em]
   
     *DeeplabV3+ \cite{DeeplabV3P} & 89.8 & 78.0  & 74.0 \\[0.8em]
     \parbox{4.8cm}{*PSPNet \cite{PSPNet}} & 89.9 & 78.0  & 74.1 \\[1.2em]
    
    \parbox{4.8cm}{*Wider Resnet-38, DeeplabV3 Decoder, Inplace ABN \cite{inplace}, Ensemble of 4} & 89.7 & 77.9  & 74.3 \\
    \bottomrule \\[-1.5ex]
  \end{tabular}
  \caption{\label{tab:sem-bench} The mIoU scores of models at 3 level of the hierarchy. The performance numbers of * models are obtained from the submissions of a AutoNUE challenge \cite{autonue} conducted based on the dataset.}
  \vspace{-1.3em}
\end{table}

    
   
    

\subsection{Class IoUs and Confusion Matrix}
\begin{figure}
    \centering
    \includegraphics[width=0.48\textwidth]{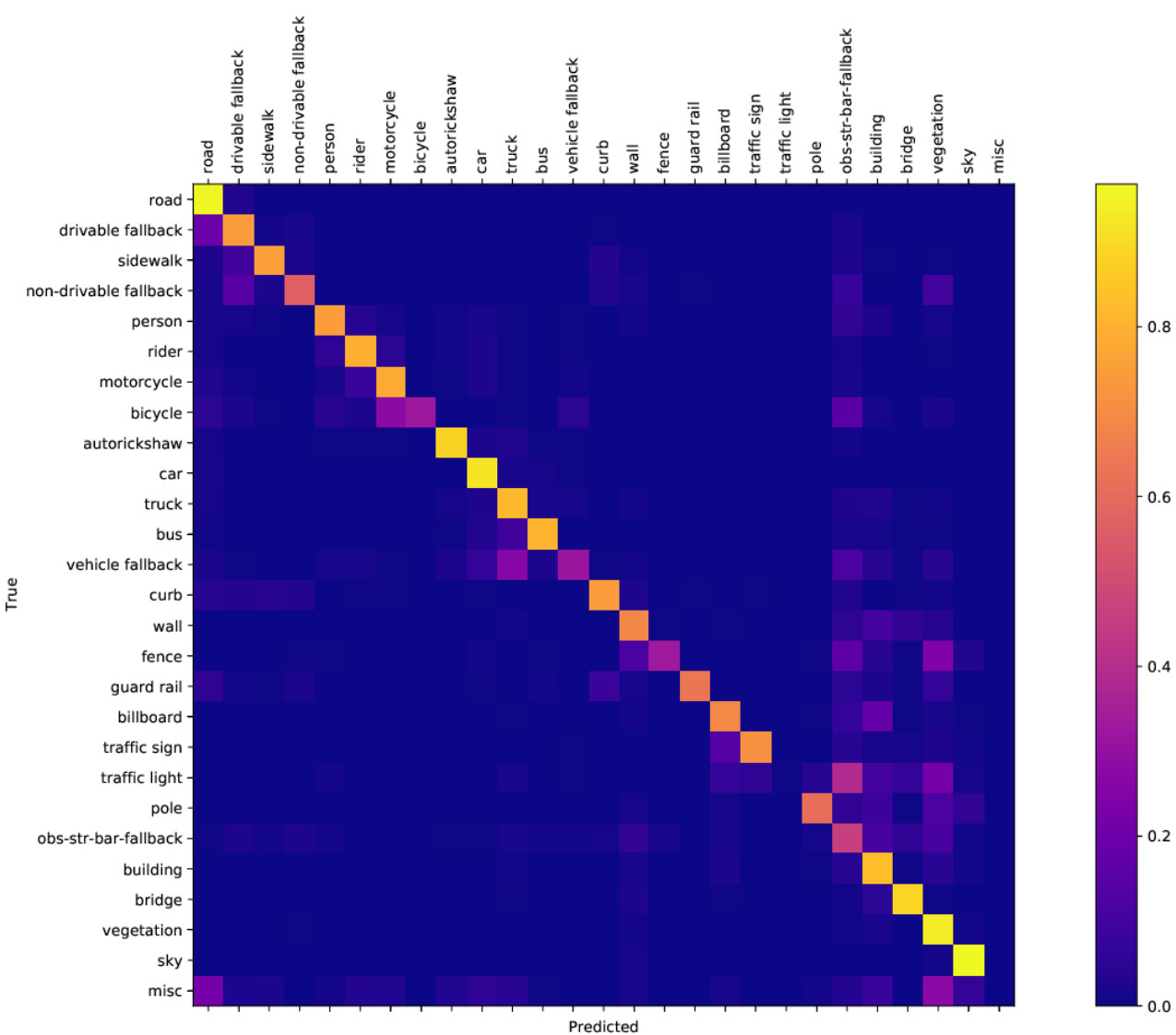}
    \caption{\label{fig:confusion} The confusion matrix of the trained model.}
   \vspace{-0.2em}
    
\end{figure}

The IoUs for every class can be found in Figure \ref{fig:ious}. We observe that the IoUs are lower than 25\% for bicycle, traffic light, vehicle fall-back and fence labels. The low scores for bicycle and traffic light can be explained by the low pixel counts. 
We also plot the confusion matrix between labels in Figure \ref{fig:confusion}. Note that there is significant confusion between:
\begin{tight_itemize}
\item motorcycle and bicycle.
\item billboard and traffic sign.
\item obs-str-bar-fallback, vegetation and traffic light.
\item building and billboard.
\item vegetation and wall, pole, fence.
\item drivable, non-drivable, vegetation.
\end{tight_itemize}

\begin{figure*}
\centering
\includegraphics[width=\textwidth]{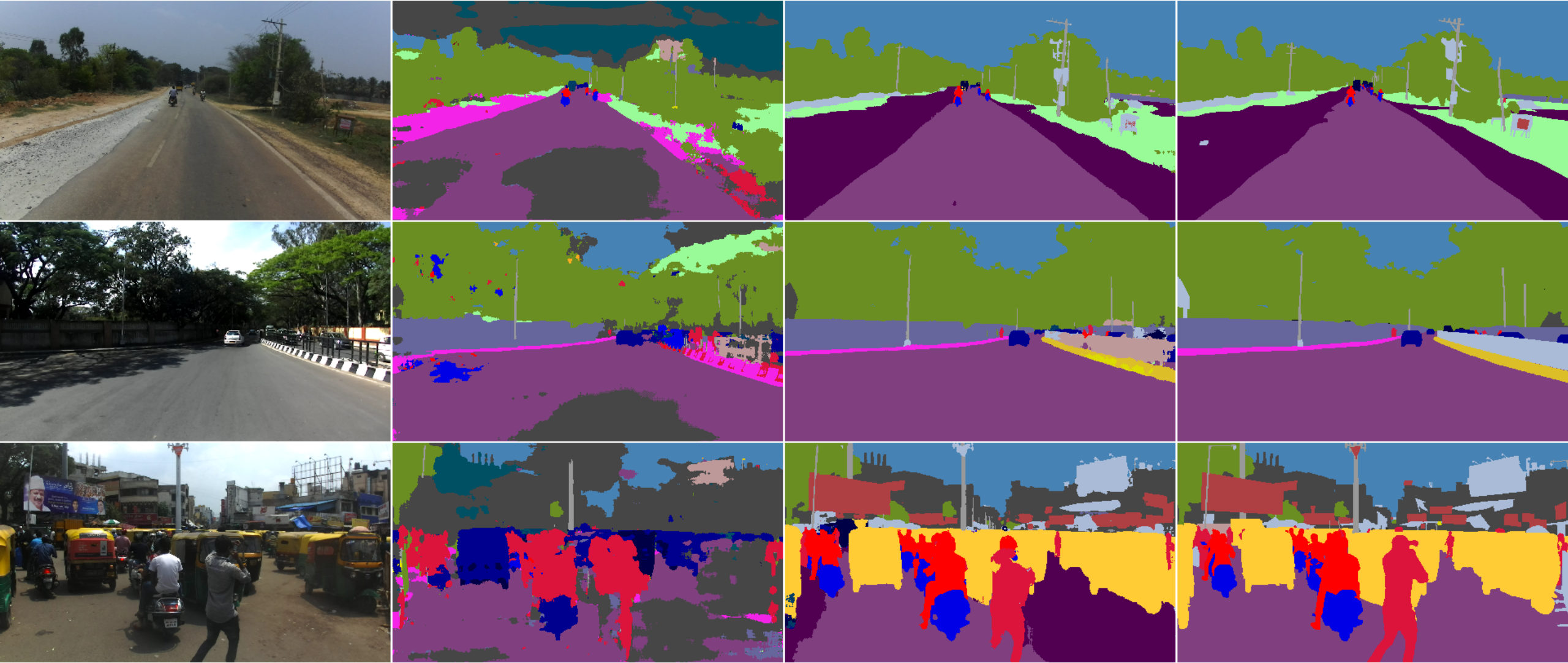}
\includegraphics[width=\textwidth]{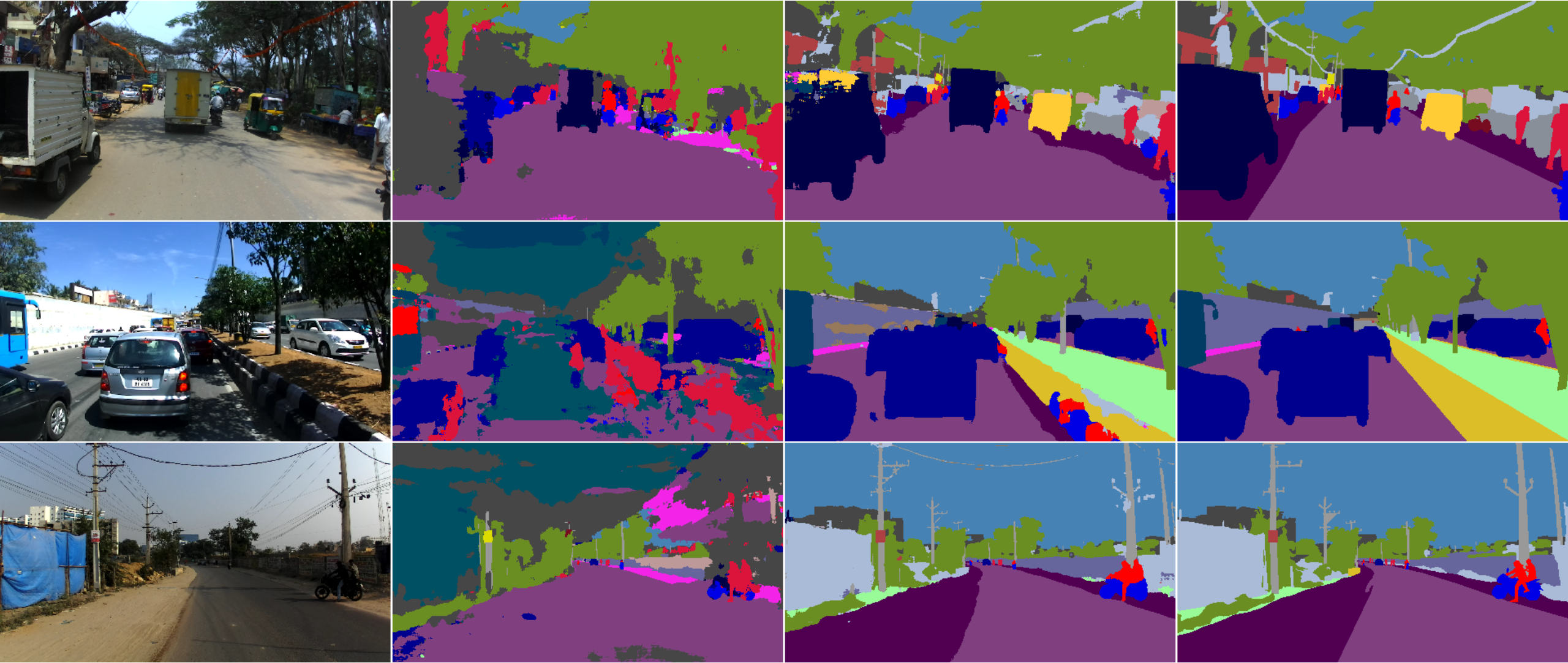}
\includegraphics[width=\textwidth]{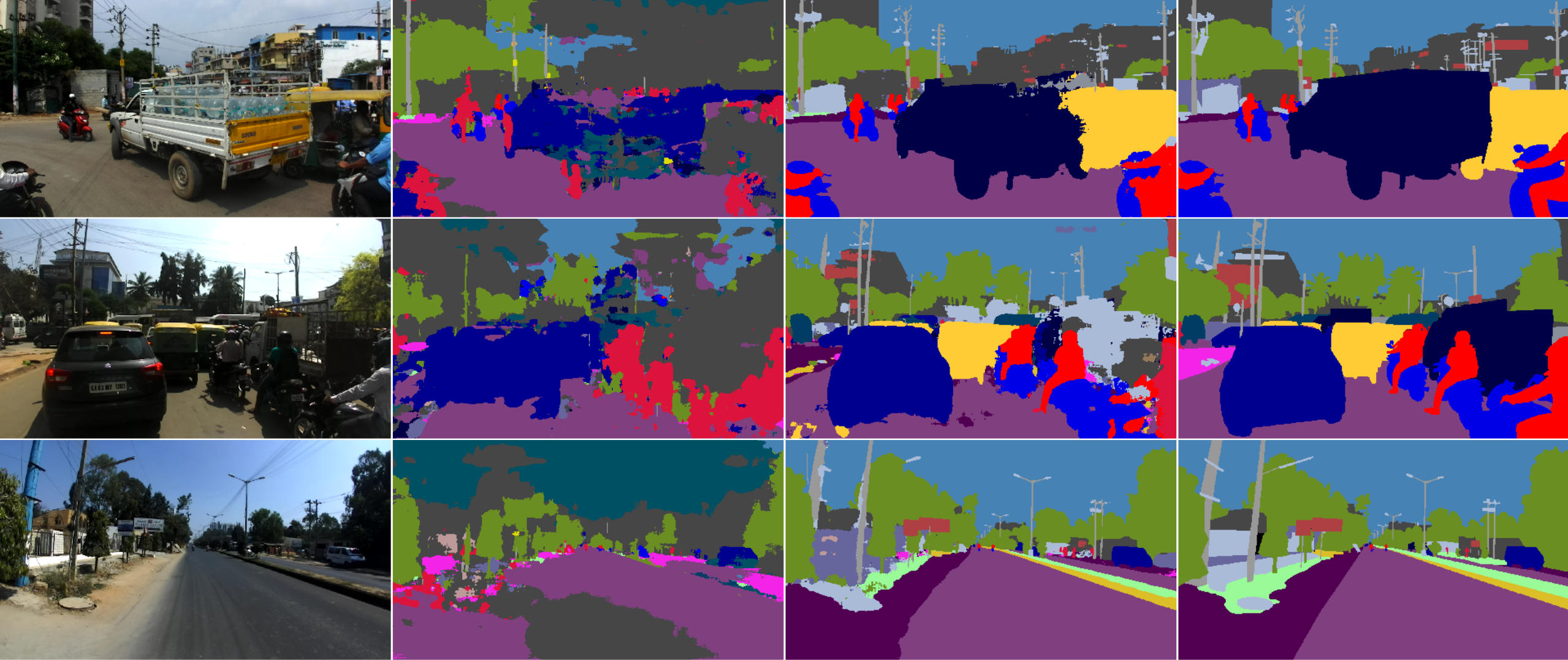}

\caption{\label{fig:qualitative} We give many qualitative example with: input image from validation set, predictions from Cityscape pretrained model, prediction from model trained on our training dataset and the ground truth in our dataset in the order of columns.}
\end{figure*}

We analyze some examples of predictions in Figure \ref{fig:qualitative}. As can be seen, model trained on our dataset gives prediction of much better quality in unstructured setting. It identifies the muddy areas which can be driven. New labels like autorickshaw, curb, billboard etc are getting identified.

\begin{table}
  \begin{tabular}{l l l l l l l}
  \toprule
    Method
    & AP &  AP@50\\
    \midrule \\[-1.5ex]
    \parbox{5cm}{*MaskRCNN \cite{MaskRCNN} with ResNet101} & 0.268 & 0.499\\[1em]
    \parbox{4.5cm}{*PANet \cite{PANet}} & 0.376 & 0.661 \\
    \bottomrule \\[-1.5ex]
  \end{tabular}
  \caption{\label{tab:inst-bench} The AP scores of models for instance labels. The performance numbers of * models are obtained from the submissions of the AutoNUE challenge \cite{autonue} conducted based on the dataset.}
  \vspace{-1.2em}
\end{table}

\subsection{Instance Segmentation Benchmark}
Similar to other datasets, we also specified an instance segmentation benchmark, where individual instances of the same label need to be segmented separately. The algorithms are required to predict a set of detections of traffic participants in the frame,  with a confidence score and a per-instance binary segmentation mask.  To assess  the
performance, the average precision on the region level for each class and average it across a range of overlap thresholds ranging  from $0.5$ to $0.95$ in  steps  of  $0.05$, similar to \cite{coco}. 

The results of some best performing submissions from the challenge are given  in Table \ref{tab:inst-bench}.

\section{Conclusion}
We present a novel dataset for studying problems of autonomous navigations in unstructured driving conditions. We identify several drawbacks of existing datasets, such as distinguishing safe or unsafe drivable areas beside the road, additional labels required for vehicles and a label hierarchy that reduces ambiguity. We analyze the label statistics and the class imbalance present in the dataset. We also examine the domain discrepancy properties with respect to other semantic segmentation datasets. \MC{In contrast to existing datasets on semantic segmentation, ours is acquired in India, which leads to greater diversity due to variations in appearance of traffic participants as well as background categories. Not only does this pose interesting challenges for the state-of-the-art in semantic segmentation, it is also the first effort in our knowledge to focus on problems related to autonomous driving in geographies outside North America or Europe with relatively less developed road infrastructure.}

\MC{In the future, we plan to extend the benchmark to computer vision problems beyond semantic segmentation. In particular, the unconstrained nature of the dataset provides a uniquely novel setting for higher-level reasoning problems such as scene understanding \cite{su1, su2, su3} and path planning \cite{pp}. Motions in the dataset are less constrained due to greater freedom in traffic participant behavior and less adherence to traffic rules. The possible absence of visual cues such as lanes that constrain traffic participant behavior  poses further challenges. Besides the presence of rare categories \cite{rare}, even common categories have diverse attributes or appearance variations. Besides, the ambient conditions differ greatly across weather, time of day and air quality. This also motivates interesting new problems for few shot learning \cite{os} and domain adaptation \cite{da}, which our future work will study in greater detail.}

{ \small
\textbf{Acknowledgements.} 
The authors would like to thank Intel, specially the Intel India
team for the efforts in capture of the data and coordination. Authors would like to specially acknowledge the support and
helps from Bharat Kaul, Prabhavathy Adavikolanu and Silar Shaik
in making this possible.
We would also like to thank Governments of Telangana and
Karnataka for the permissions, encouragement and enabling
this effort.
}
\pagebreak
{
\bibliographystyle{ieee}
\bibliography{egbib}
}

\end{document}